\documentclass[conference]{IEEEtran}
\IEEEoverridecommandlockouts
\usepackage{cite}
\usepackage{amsmath,amssymb,amsfonts}
\usepackage{algorithmic}
\usepackage{textcomp}
\usepackage{xcolor}
\usepackage{subcaption}
\usepackage{array}
\usepackage{graphicx}
\graphicspath{{./Chapters/Figures/}}
\usepackage[export]{adjustbox}
\usepackage{tikz}
\usepackage{float}
\usetikzlibrary{arrows}
\usepackage{verbatim}
\usepackage{bibentry}
\usepackage{graphicx}
\graphicspath{Chapters/Figures}
\def\BibTeX{{\rm B\kern-.05em{\sc i\kern-.025em b}\kern-.08em
		T\kern-.1667em\lower.7ex\hbox{E}\kern-.125emX}}
\begin{document}
\title{Machine Learning and Data Science approach towards trend and predictors analysis of CDC Mortality Data for the USA
    }
	\author{
		\IEEEauthorblockN{Yasir Nadeem}
		\IEEEauthorblockA{\textit{Department of Computer Science} \\
			\textit{Mohammad Ali Jinnah University}\\
			Karachi, Pakistan \\
			fa17bscs0028@maju.edu.pk}
		\and
		\IEEEauthorblockN{Awais Ahmed}
		\IEEEauthorblockA{\textit{Department of Computer Science} \\
			\textit{Mohammad Ali Jinnah University}\\
			Karachi, Pakistan \\
			awais.ahmed@jinnah.edu}
	}
\maketitle
\begin{abstract}
The research on mortality is an active area of research for any country where the conclusions are driven from the provided data and conditions. The domain knowledge is an essential but not a mandatory skill (though some knowledge is still required) in order to derive conclusions based on data intuition using machine learning and data science practices. The purpose of conducting this project was to derive conclusions based on the statistics from the provided dataset and predict label(s) of the dataset using supervised or unsupervised learning algorithms. The study concluded (based on a sample) life expectancy regardless of gender, and their central tendencies; Marital status of the people also affected how frequent deaths were for each of them. The study also helped in finding out that due to more categorical and numerical data, anomaly detection or under-sampling could be a viable solution since there are possibilities of more class labels than the other(s). The study shows that machine learning predictions aren't as viable for the data as it might be apparent.

\end{abstract}
	
\begin{IEEEkeywords}
Mortality, Machine learning, Trends, Predictors.
\end{IEEEkeywords}

\section{Introduction}
The analysis of data is important to businesses and (in this case) to scientists as it uncovers and helps them better understand the statistical facts and figures which in practice, enables them to extract meaningful information and to draw qualitative or quantitative conclusions. The analysis for mortality is done to identify the fatality causing problems that would help the researchers to work towards the solution. The mortality statistics of any country are arguably an important aspect that, in some cases, depicts the state of the country or a region, how satisfied the people are and how advanced the medical facilities probably are.

The rapid advancement in tools and techniques gave birth to Machine learning and Data Science and led them to be available to researchers to further the advancement, especially in the medical field, which has been a trend since its inception. However, according to the research and based on the research papers, very few mortality based trends and predictors analysis is done using the aforementioned tools and techniques. Therefore, it was motivating for us to carry out analysis in this area in order to help other researchers gain a ground upon which to base further conclusions.

The plan to carry out the research was to acquire the dataset(s) from an authentic and reputable source to obtain factual data upon which the research will be based and formulate questions to which answers will be searched after which the data will then be transformed, cleaned and meaningful information would be extracted through data science practices and finally, follow the machine learning route and apply them to predict and learn patterns that would advance the study towards the end. The study is an extension of the semester project which was to be completed under the period of half a semester under the supervision of a professor.

This paper is intended to follow the pattern where: Section 2 provides the literature review. Section 3 would describe the methodology along with what dataset I used and the proposed approach concerning the title. Finally, the Evaluation metrics and results are shown in Section 4 and 5 respectively. The study is then concluded in Section 6.

\section{LITERATURE REVIEW}
The literature for this area of study is countable as in "few", however, they are fairly recent which means this area of research became active as the tools and techniques suitable to them were made available. The researches in this area are more specific compared to this study, however, they bring an interesting point of view towards the problems they outlined and intended to solve.

The analysis of the CDC mortality data for 2003 till 2013 showed that the data science approach could be applied to 18 variables selected from the study to predict the probability of whether the sample's data counts towards suicide or not. The study while referencing the breakthrough of data science when applied to genome sequencing also outlines the potential use of machine learning algorithms such as Clustering, which is an unsupervised learning algorithm to dataset(s) to learn patterns and group certain samples into the same group (cluster) based on similarity; Regression which is a supervised learning algorithm of two types: Linear, where a linear line is fitted (after learning) to the data or Logistic, where a decision boundary is created to differentiate between the labeled data. \cite{diaz2017translating}

A deep learning model based on Convolutional Neural Network was developed and used to model real-time all cause mortality in pediatric intensive care unit in South Korea from January 2011 to December 2017 that enabled them to predict mortality of patients at high risk for mortality with upto 60 h prior to death having an accuracy of 80\% with 7\% false alarms. \cite{kim2019deep}

A super-learning i.e. an ensembles approach, allowing them to use multiple algorithms to do a comparative analysis between different machine learning algorithms. The research showed that according to the data of the subjects in California, USA, there were 269 deaths within 5 years of baseline. The majority of subjects (59\%) were female. The largest age group was $ 70 \leq x \leq 80 $ years where 50\% of the subjects had self-rated health as "good", 32\% with "excellent", and 15\% with "fair" in March 2013. \cite{rose2013mortality}

\section{METHODOLOGY}
\subsection{Dataset:} For the analysis it was important to pick the a dataset from an authentic source upon which we could confidently base this study on. The latest, which at the time of writing is 2017, Mortality Multiple Cause Files dataset was picked from "Centers for Disease Control" (CDC) upon which our main analysis points are based though a 2007 dataset from the same category was picked in order to compare statistics and prediction results between the two.

The dataset had privacy as its concern for ethical considerations, therefore, some of the features of the data were encoded or removed from the actual dataset and were published on their data center. The recoded feature names, however, were detailed in the provided user guide along the dataset which we utilized extensively in order to derive conclusions from the dataset. The dataset in total contains 2846305 samples and 75 feature vectors, however, due to computational power unavailability and time limitations, only 100k sample points were used for the study.

\begin{center}
\begin{tabular}{ c c c }
\hline
 Resident Status & Age\_Value & Age\_Recode \\ 
 \hline
 1 & 71 & 40 \\ 
 1 & 74 & 37 \\
 3 & 59 & 40\\ 
 3 & 90 & 44\\ 
 2 & 66 & 39\\ 
 1 & 84 & 42\\
 3 & 88 & 38\\ 
 1 & 53 & 47
\end{tabular}
\\
\vspace{1mm}
Table 1: Dataset Example
\end{center}

\subsection{Data Loading:} The loading process involved processing the data in its raw format and transform it into a convenient format such as "CSV" to make it useable and readable by the popular programs and tools. The data was processed (raw format) using a python script \cite{diaz2017translating} that would read the raw dataset file and write it to a different CSV file. Both the dataset for 2018 and 2007 were fed into the script to obtain two separate CSV files that will be used further in the study.

\subsection{PreProcessing:} Pre-processing is the most important part of a data scientist's standard operating procedures. The pre-processing is emphasized because the data that we might need to work on will not always be in the form that we could apply analysis on effectively. The data might not even be in a structured form i.e. unstructured or raw text or format. In this case, it was in raw format which is stated in the previous section. Pre-processing also includes fixing the data by eliminating or mitigate the faults, defects, and missing information using data science techniques to make it "fit for purpose".

Therefore, the next chronological step, after loading the data, is to pre-process the data. In our case, pre-processing the data included the data wrangling steps i.e. Data cleaning, normalizing, imputation or replacement with mean (if real numbers) or mode (if numerical/categorical) and last but not the least, outlier detection and elimination. One of the most important decisions was to how much values to keep in the dataset for analysis as some of the columns had few non-missing values. Therefore, after trial and error, the threshold was decided to be kept at 30\% non-missing values which in theory means that we have 70\% of the data missing which we had to deal with. One of the methods we tried was to replace the other values with the mode/mean of the values but this would result in very skewed and disastrous results later in the pipeline, especially when using the data for training and testing for machine learning. Therefore, another reason for the aforementioned pre-processing steps is because it conforms well to a concept in machine learning i.e. feature engineering.

The next stage was to analyze the values or rather the types of the columns of the dataset. The dataset had few blank columns such as Education column (shown in Table 2) and few Entity axis conditions (EAC) encodes which had blanks as well as $<$30\% of values populated in them, some of the Record axis conditions (RAC) columns were also blank with some of them being in the same pool as EAC features. It turns out that after applying the pre-processing techniques, from the total 75 columns, only 34 columns ended up being usable from which 21 were "numerical" and 13 were found to be in ordinal or non-ordinal "string" format.

\begin{center}
\begin{tabular}{ c c c }
\hline
 Manner\_Of\_Death & Sex & Education \\ 
 \hline
 7 & F & NaN \\ 
 7 & M & NaN \\
 3 & F & NaN\\ 
 3 & M & NaN\\ 
 2 & F & NaN\\ 
 7 & M & NaN\\
 6 & F & NaN\\ 
 7 & M & NaN
\end{tabular}
\\
\vspace{1mm}
Table 2: Missing Data Example
\end{center}

Since the data had fields that were missing altogether from the dataset for example "Education" as evident in Table 2. For the aforementioned features, there was no choice but to impute the whole column as they cannot, in any circumstance, be used for analysis since they are blank and will not contribute towards the analysis and doing so we do not lose the other information.

Through these steps the data was then prepared to be processed for further descriptive and exploratory data analysis.

\subsection{Insights} This section talks about gaining insights from the data after the data is pre-processed and is considered to be the next chronological step. Gaining insights from the data is an important task for a data scientist, after all, we are performing analysis and if we do not get answers for our questions then it entails the failure of effective analysis and rather fulfills the task of analysis as a formality.

The insights was also one of the goals involved while planning the research plan and was part of the to-do list for the semester project. Some of the insights hence, involved finding out:

\begin{itemize}
    \item Trends in mortality data-set i.e. how causes of deaths have changed when compared to past.
    \item If married people die more than unmarried regardless of gender.
    \item If married male die more than females.
    \item If unmarried female die more than married females.
    \item the marital status of the most common deaths by comparing death rates.
    \item Age distribution for the deaths.
    \item Life expectancy for certain age groups.
    \item Whether people died more while not at work.
    \item The most common cause of death.
    \item Whether cause of deaths changed over last 11 years.
\end{itemize}

\subsection{Descriptive Data Analysis:}
This section involves applying the statistical knowledge on the dataset to find factual data that would later support our study or rather quantify it. This step is also known as "Statistical analysis" as the previous description might suggest. The statistical analysis in this study in involved finding out basic statistical measures of each or related feature vectors such as mean, median, mode, standard deviation, variance, and, inter-quartile range(s). The statistical measures were also taken for the insights that required statistical measures and for detection and elimination of outliers.

\[ Mean = \frac{1}{N}\sum\limits_{i=1}^{N}{x_{(i)}} \]
\[ Median = \left(\frac{N+1}{2}\right)^{th} \textit{(If N is Odd)} \]
\[ Median = \frac{1}{2}\left(\left(\frac{N}{2}\right)^{th} + \left(\frac{N+2}{2}\right)^{th}\right) \textit{(If N is Even)} \]
\[ Mode = l + h\left(\frac{f_m-f_1}{2f_m-f_1-f_2}\right) \]
\[ Standard Deviation = \sigma = \sqrt{\frac{\sum\limits_{i=1}^{N}{(X_{i} - \mu)}}{N}} \]
\[ Variance = \sigma^2 = \frac{\sum\limits_{i=1}^{N}{(X_{i} - \mu)}}{N} \]
\[ IQR = \frac{3}{4}\left( N+1 \right)^{th} - \frac{1}{4}\left( N+1 \right)^{th} \]

\subsection{Exploratory Data Analysis} Exploratory data analysis, as the name suggests, implies that we have to explore the data for the insights, to find meaningful information and visualize them in order to understand the data better and to explain it better to others. Therefore, in this section the dependencies were looked at by correlating multiple features using correlation line graphs in order to visualize the correlations, comparing categories of feature vectors, visualizing data against our insights which would finally conclude the EDA.

In this section of analysis, the questions that were formulated as a part of the semester project after pre-processing were then answered by using the feature vectors and gathering statistics from them. The visualizations can also be called "the essence" of EDA, therefore, the graphs were made according to the insights and dependencies of the feature vector. For example: A histogram will be made if distribution of a feature vector is to be found, however, if correlations are to be found of two independent variables that are dependent on a single feature then a line graph is plotted for both variables over the dependent variable.

\subsection{Machine Learning Data Split} The dataset had undergone many trial and error cycles whose analysis yielded different features that could be used as a target for a subset of features such that: 

\[ F \subset D \]
\[ T \subset D \]
\[ T \not\subset F \]

Here, "F" represent features, D represents the Dataset and T represents the Target feature vector. The data was then split into 80\% and 20\% ratio for training and testing respectively.

\section{Evaluation Measures}
The evaluation of machine learning models in an important task as it allows us to measure how well the performance of algorithms is. The evaluation to make more sense needs to be in a format that we can quantify and effectively use in calculations entailing the need for real numbers in performance metrics. It also allows us to compare the performance of different algorithms based on quantification.

The evaluation in machine learning is done after the machine learning algorithm is applied to the dataset, which is generally known as "training" in the machine learning community. Therefore, evaluation measure in our case was taken after the model(s) were trained, however, due to the short time the measures were taken after testing. The evaluation measures used accuracy score along with $ R^2 $ score to evaluate the score for testing data and how accurately the model predicted the labels for numerical or categorical data or values if the target to be predicted was a real number respectively.

\[ Accuracy\  Score = \frac{correct\  predictions}{number\  of\  samples} \]

\[ R^2 = 
\frac{n.\sum{xy} - (\sum(x).\sum(y))}
{\sqrt{[n.\sum{x^2} - (\sum{x})^2].[n.\sum{y^2} - (\sum{y})^2]}}
\]

\section{RESULTS AND DISCUSSION} Perhaps the most exciting and awaited part of any study is the results along with discussion. The result, in this case, aims to show the findings that were obtained after multiple stages of processes. The result also aims to be intuitive to the reader in a sense that the results, at least the insights, are visual. Therefore, the results section is divided into further subsections that will address their respective objectives. 

\subsection{Data Intuition}

Intuition can be defined as "The ability to acquire knowledge without proof, evidence, or conscious reasoning, or without understanding how the knowledge was acquired." The intuition however wrong, sometimes is a necessary evil that helps us to move in the right direction. In our study, the data at first was assumed to be flawless after pre-processing and that the data provided necessarily had all the details needed in order to derive conclusions from it or to be used in order to correlate depenedent or independent feature vectors. The results, however, showed that intuition was right in some case but most of the times, for the data, it became ever so evident.

\subsection{Visual Insights}
This section provides us with the ability to better judge the data, visualize it and show our findings for the insights that we initially expected from the study. Each figure addresses a subset of question or rather insight that were formulated after pre-processing of the data.

The following plot shows the general age distribution of the samples in the dataset. In the plot, we can see that the distribution is left-skewed which means that the mode is greater than the median which further greater than the mean of the ages. The distribution is plotted without any regard for gender. Here, the Mode is 83, the median is 75 and the mean is 72.

\includegraphics[width=0.486\textwidth]{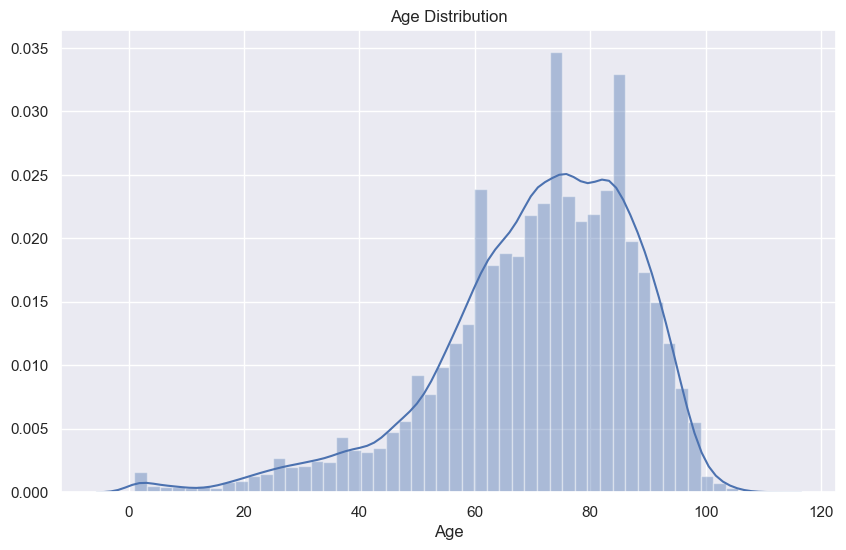}

The following graph shows the age distribution for the suicides data, the sample size for the suicides is smaller than the general age distribution because interval for each histogram bar is smaller than the histogram for general age distribution. This distribution has two peaks i.e. with the main peak at 27 and the low peak around 62. This distribution is also known as "bimodal" distribution among the statisticians.

\includegraphics[width=0.486\textwidth]{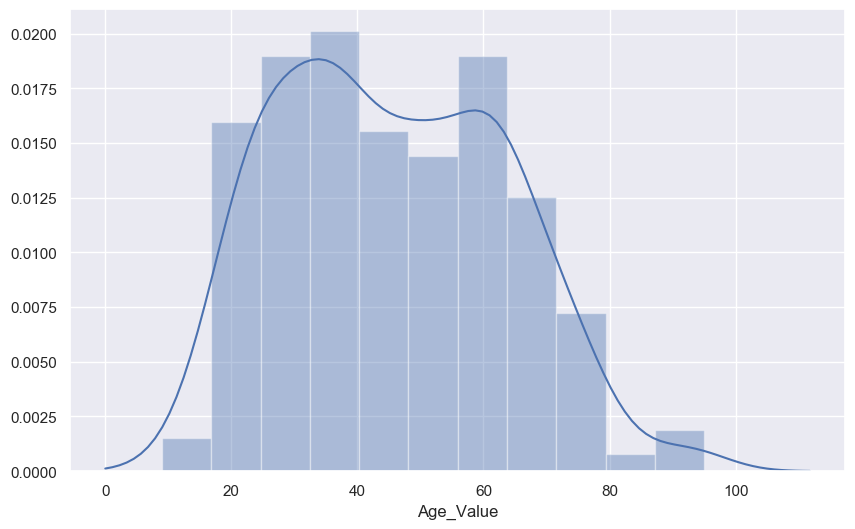}

The following graph shows the life expectancy for the deceased according to the data. Upon closer inspection, it reveals that the samples, without regard to the gender and looking at the highest peak, are more likely to die at the age interval of 70-79.

\includegraphics[width=0.486\textwidth]{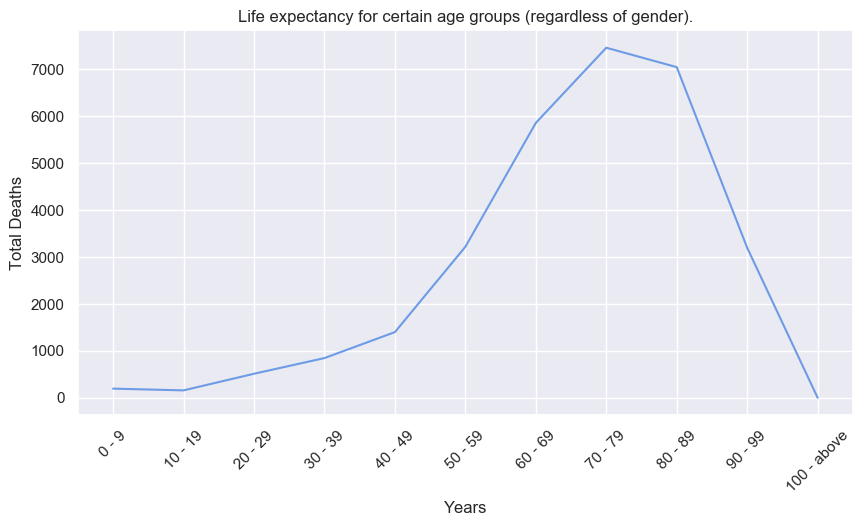}

The following graph shows the comparison between male and female life expectancy or deaths. According to the plots, male are more likely to die around the interval of 70-79 and females are more likely to die in the interval of 80-89. The previous general sample foreshadowed this result with it's double peak in line plot. Therefore, the 100k sample point shows that females are more likely to live longer than the males.

\includegraphics[width=0.486\textwidth]{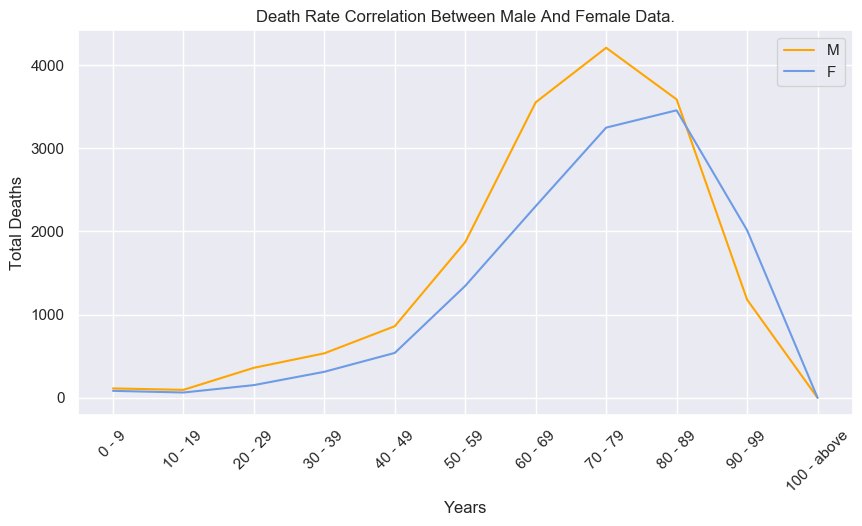}

The following graph shows the 2018 death cause of the pie plot. The plot shows that natural deaths are more likely at around 91\% of the time when compared to accidents at around 7\% of the time while other causes are 1-2\% representative of total samples death. The currently investigating cause of death has been excluded as it wasn't significant enough to include.

\includegraphics[width=0.485\textwidth]{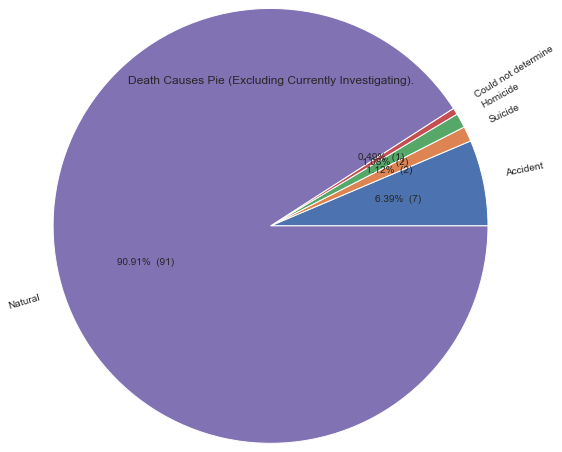}

The following graph shows the 2018 vs 2007 death comparison for the deceased according to the data.

\includegraphics[width=0.486\textwidth]{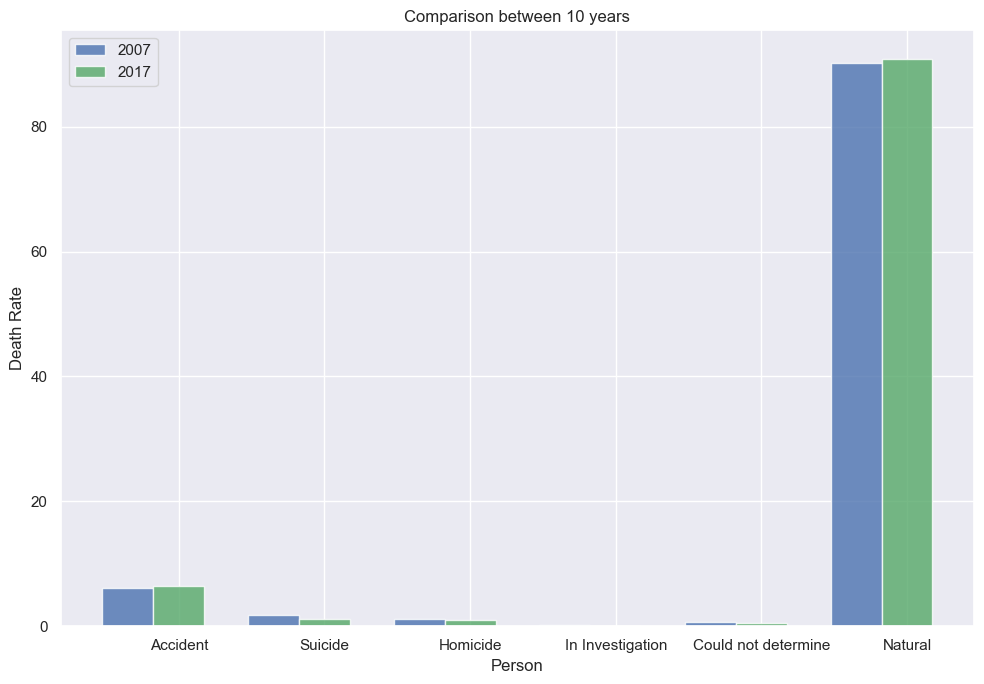}

The following graph shows the people who died at work. According to the chart it is evident that most of the deaths aren't caused at death. The measures show that 0.16\% of the people die at work, around 8.8\% of deaths were caused while not at work, whereas, around 91\% of the sample's death location was unknown.
\begin{center}
    \includegraphics[width=0.45\textwidth,height=0.32\textheight]{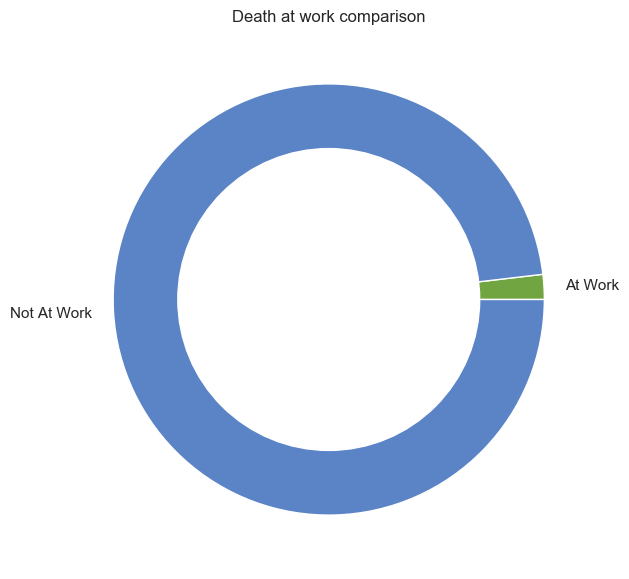}
\end{center}

The following graph shows the most lethal disease according to 100k data point sample. Here, J449 occupies more than 50\% of the samples while C349 occupies a little over 50\% of the samples while others occupy less than 50\% of the samples.\newline
\includegraphics[width=0.485\textwidth]{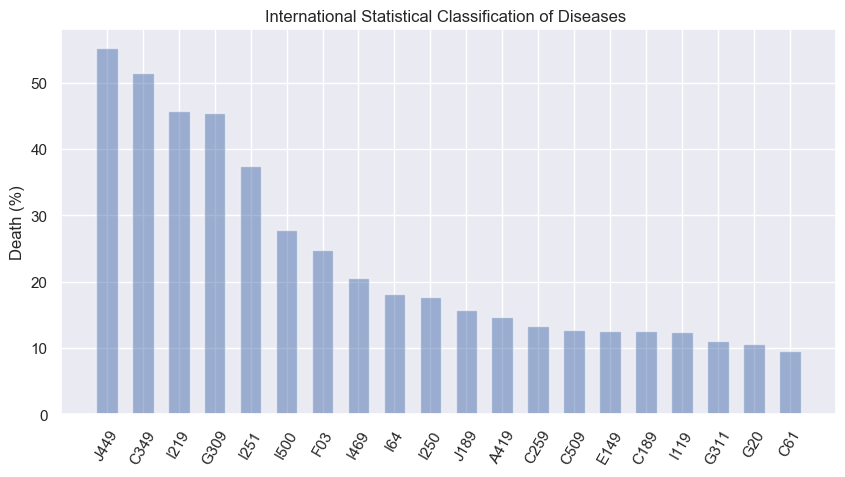}

According to the data, for 100k data points, J449 is related to \textbf{Chronic obstructive pulmonary disease and Skin Cancer} and C349 was the second most common disease of death among the deceased and is related to \textbf{Malignant neoplasm of bronchus and lung cancer}. [4]

\subsection{Problems}
The ML approach yielded that we could predict marital status 66\% of the time but it can be attributed to the fact that there was a class imbalance present for marital status as well though less obvious than the other feature vectors for which it was almost impossible due to either the fact that our pre-processing replacement had flaws or the data provided was flawed which eventually caused the imbalance in class labels, meaning there was a class label that occurred more than the others, therefore high accuracy in model implied classes imbalances rather than accurate predictions. In order to utilize data in this form, Anomaly detection was found to be the solution. This would allow us to detect false information or events in the data were they to occur in the data in future. The second option, due to class imbalance, would be to under sample the data and use it to predict and analyze the patterns.

\subsection{Limitations}
Although the research yielded meaningful information and aided us in answering the questions from the insights, more could have been found if not for the limitations in the study. The biggest limitation in the study was time since we had to complete the research in half a semester which equates to 2 months. The second limitation was the knowledge of how the processes work since this was part of an introductory course and we did not have information to work with the data beforehand, therefore, applying while learning was deemed to be the viable solution in the study. The last but not the least limitation was the domain knowledge even with access to the professionals who have domain knowledge might have led to better results but it again leads to the first limitation which is "time".

\section{Conclusion}
The study revealed answers to many interesting questions that were generated during the insights stage and that it would prove useful to the other researchers searching for latest analysis or review on the dataset used in the study. The possibilities and impossibilities analysis also helped us find solutions to the class imbalance problems according to which anomaly detection or under-sampling could be the solution. The machine learning approaches explored with the data did not yield result, granted with our methodology and hence, did not prove to be beneficial. We dully hope this study would prove to be fruitful for future researchers as to what other methodology could be followed and how further studies are to be done.
\clearpage
\nocite{*}
\bibliographystyle{unsrt}
\bibliography{r} 
\end{document}